\documentclass[runningheads,a4paper]{llncs}
\usepackage{amssymb}
\usepackage{graphicx}
\usepackage{algorithm}
\usepackage{algorithmic}
\usepackage[lined,algonl,boxed,algo2e]{algorithm2e}
\usepackage{caption}
\usepackage{subcaption}
\usepackage{url}
\newcommand{\keywords}[1]{\par\addvspace\baselineskip
\noindent\keywordname\enspace\ignorespaces#1}
\usepackage{amsmath}
\usepackage{amssymb}
\usepackage[colorlinks]{hyperref}
\usepackage{lipsum}

\usepackage{xcolor}
\hypersetup{colorlinks, linkcolor={red!50!black}, citecolor={red!50!black},  urlcolor={red!50!black}}
\usepackage[switch*]{lineno}
\begin{document}

\mainmatter  % start of an individual contribution

\title{A Visual Quality Index for Fuzzy C-Means}
\author{Ayb\"uke \"Ozt\"urk
\and  St\'ephane Lallich
\and J\'er\^{o}me Darmont}
\institute{{Universit\'e de Lyon, Lyon 2, ERIC EA 3083\\
5 avenue Pierre Mend\`es France, F69676 Bron Cedex, France \\
aybuke.ozturk@univ-lyon2.fr, stephane.lallich@univ-lyon2.fr, jerome.darmont@univ-lyon2.fr
} \\ 
}

\maketitle

\begin{abstract}

Cluster analysis is widely used in the areas of machine learning and data mining. Fuzzy clustering is a particular method that considers that a data point can belong to more than one cluster. Fuzzy clustering helps obtain flexible clusters, as needed in such applications as text categorization. The performance of a clustering algorithm critically depends on the number of clusters, and estimating the optimal number of clusters is a challenging task. Quality indices help estimate the optimal number of clusters. However, there is no quality index that can obtain an accurate number of clusters for different datasets. Thence, in this paper, we propose a new cluster quality index associated with a visual, graph-based solution that helps choose the optimal number of clusters in fuzzy partitions. Moreover, we validate our theoretical results through extensive comparison experiments against state-of-the-art quality indices on a variety of numerical real-world and artificial datasets. 

\end{abstract}

% INTRODUCION
\section{Introduction}

Clustering refers to the assignment of unlabeled data points into clusters (groups) so that the points belonging to the same cluster are more similar to each other than those within different clusters. There are various types of clustering strategies, including crisp and fuzzy clustering. In crisp (or hard) clustering, a data point can belong to one and only one cluster, while in fuzzy clustering \cite{ruspini1970numerical}, a data point can belong to several clusters. Fuzzy clustering is very useful in many applications, e.g., the text categorization of various news into different clusters: a science, a business, and a sport cluster; where an article containing the keyword "gold" could belong to all three clusters. Furthermore, it is also possible to open discussions with domain experts when using fuzzy clustering. 

Clustering algorithms behave differently for different reasons. The first reason relates to dataset features such as geometry and the density distribution of clusters. The second reason is the choice of input parameters such as the fuzziness coefficient $m$ ($m = 1$ indicating that clustering is crisp and $m > 1$ that clustering becomes fuzzy).

These parameters all affect the quality of clustering. To study how the choice of parameters impacts clustering quality, we need a quality criterion. For instance, when the dataset is well separated and has only two variables, a scatter plot can help determine the number of clusters. However, when the dataset has more than two variables, a good quality index is needed to compare various cluster configurations and choose the appropriate number of clusters.

Achieving a good clustering involves both minimizing intra-cluster distance (compactness)  and maximizing inter-cluster distance (separability). A common issue in this process is that clusters are split up while they could be more compact. Many cluster quality indices have been proposed to address this problem for hard and fuzzy clustering, but none of them is always highly efficient~\cite{pal1997correction}. 

Moreover, there is no real-life golden standard for clustering analysis, since various experts may have different points of views about the same data and express different constraints on the number and size of clusters. Thanks to a visual index, different solutions can be presented with respect to the data. Thus, experts can make a trade-off between their opinion and the best local solutions proposed by the visual index.

Hence, in this paper, we first review existing quality indices that are well-suited to fuzzy clustering, such as~\cite{bezdek1973cluster,chen2001rule,calinski1974dendrite,fukuyama1989new,xie1991validity,zhang2014novel}. Then, we propose an innovative, visual quality index for the well-known Fuzzy C-Means (FCM) method. Moreover, we compare our proposal with state-of-the-art quality indices from the literature on several numerical real-world and artificial datasets.

The remainder of this paper is organized as follows. Section~\ref{sec:PrincipleofFuzzyClustering} recalls the principles of fuzzy clustering. Section~\ref{sec:RelatedWorks} surveys quality indices for fuzzy clustering. Section~\ref{sec:AnIndexAssociatedwithVisualSolution} details our visual quality index. Section~\ref{sec:Experiments} reports on the experimental comparison of our quality index  against existing ones on different datasets. Finally, we conclude this paper and provide research perspectives in Section~\ref{sec:Conclusion}. 

\section{Principles of Fuzzy Clustering}
\label{sec:PrincipleofFuzzyClustering}

Fuzzy inertia is a core measure in fuzzy clustering. Fuzzy inertia $FI$ (Equation~\ref{eqFI}) is composed of fuzzy within-inertia $FW$ (Equation~\ref{eqFW}) and fuzzy between-inertia $FB$ (Equation~\ref{eqFB}). Membership coefficients $u_{ik}$ of data point $i$ to cluster $k$ are usually stored in a membership matrix $U$ that is used to calculate $FW$, $FB$ and $FI$. Note that $FI$ = $FW$ + $FB$. Moreover, $FI$ is not constant because it depends on $u_{ik}$. When $FW$ changes, the values of $FI$ and $FB$ also change.

\begin{equation} \label{eqFI}
FI = \sum_{i=1}^n \sum_{k=1}^K  u_{ik}^m d^2 (x_i,\overline{x})
\end{equation}

\begin{equation} \label{eqFW}
FW = \sum_{i=1}^n \sum_{k=1}^K  u_{ik}^m d^2 (x_i,c_k)
\end{equation}

\begin{equation} \label{eqFB}
FB = \sum_{i=1}^n \sum_{k=1}^K  u_{ik}^m d^2 (c_k,\overline{x})
\end{equation}

where $n$ is the number of instances, $K$ is the number of clusters, $m$ is the fuzziness coefficient (by default, $m = 2$), $c_k$ is the center of the $k^{th}$ cluster $\forall 1 \leq k \leq K$, $\overline{x}$ is the grand mean (the arithmetic mean of all data -- Equation~\ref{Equation_arithmeticMean}), and function $d^2()$ computes the squared Euclidean distance.
        
\begin{equation} \label{Equation_arithmeticMean}
\overline{x} = \frac{1}{n}\sum_{i=1}^n x_i
\end{equation}

FCM is a common method for fuzzy clustering that adapts the principle of the K-Means algorithm \cite{macqueen1967some}. FCM, proposed by \cite{dunn1973fuzzy} and extended by \cite{bezdek1984fcm}, applies on numerical data. Since numerical data are the most common case, we choose to experiment our proposals with FCM.

The aim of the FCM algorithm is to minimize $FW$. It starts by choosing $K$ data points as initial centroids of the clusters. Then, membership matrix values $u_{ik}$ (Equation~\ref{FCMMembershipM_eqn}) are assigned to each data point in the dataset. Centroids of clusters $c_k$ are updated based on Equation~\ref{FCMCentroid_eqn} until a termination criterion is reached successfully. In FCM, this criterion can be a fixed number of iterations $t$, e.g., $t = 100$. Alternatively, a threshold $\epsilon$ can be used, e.g., $\epsilon = 0.0001$. Then, the algorithm stops when $FW_{K+1}$/ $|$ $FW_{K+1}$ - $FW_{K}$ $|$ $< \epsilon$.

\begin{equation} \label{FCMMembershipM_eqn}
u_{ik} = \frac{1}{\sum_{j=1}^K (\frac{\|x_i-c_k\|^2}{\|x_i-c_j\|^2})^\frac{1}{m-1}} 
\end{equation} 

\begin{equation} \label{FCMCentroid_eqn}
c_k = \frac{\sum_{i=1}^nu_{ik}^mx_i}{\sum_{i=1}^nu_{ik}^m}
\end{equation}

\section{Fuzzy Clustering Quality Indices}
\label{sec:RelatedWorks}

According to Wang et al.~\cite{wang2007fuzzy}, there are two groups of quality indices. Quality indices in the first group are based only on membership values. They notably include partition coefficient index $V_{PC}$ \cite{bezdek1973cluster} (Equation~\ref{PC_eqn}; $\frac{1}{K} \leq V_{PC} \leq 1$; to be maximized) and Chen and Linkens' index $V_{CL}$ \cite{chen2001rule} (Equation~\ref{ChenLinked_eqn}; $0 \leq V_{CL} \leq 1$; to be maximized). $V_{CL}$ takes into consideration both compactness (first term of $V_{CL}$) and separability (second term of $V_{CL}$).

\begin{equation} \label{PC_eqn}
V_{PC}= \frac{1}{n} \sum_{i=1}^n \sum_{k=1}^K u_{ik}^2
\end{equation} 

\begin{equation} \label{ChenLinked_eqn}
V_{CL} = \frac{1}{n} \sum_{i=1}^n max_{k}(u_{ik}) - \frac{1}{c} 
\sum_{k=1}^{K-1} \sum_{j=k+1}^K \Bigg[\frac{1}{n} \sum_{i=1}^n min(u_{ik}, u_{ij})\Bigg],
\end{equation}

where $c = \sum_{k=1}^{K-1}k $. \\

Quality indices in the second group associate membership values to cluster centers and data. They include an adaptation of the Ratio index $V_{FRatio}$ to fuzzy clustering \cite{calinski1974dendrite} (Equation~\ref{Ratio_eqn}; $0 \leq V_{FRatio} \leq +\infty$; to be maximized), Fukuyama and Sugeno's index $V_{FS}$ \cite{fukuyama1989new} (Equation~\ref{Fukuyama_eqn}; $-FI \leq V_{FS} \leq FI$; to be minimized), and Xie and Beni's index $V_{XB}$ \cite{xie1991validity,pal1995cluster} (Equation~\ref{XieBeni_eqn}; $0 \leq V_{XB} \leq FI/n*min\|x_j - v_k\|^2 $; to be minimized).
 
\begin{equation} \label{Ratio_eqn}
V_{FRatio} = FB/FW
\end{equation}

\begin{equation} \label{Fukuyama_eqn}
V_{FS} = FW - FB
\end{equation}

\begin{equation} \label{XieBeni_eqn}
V_{XB} = \frac{\sum_{k=1}^K \sum_{i=1}^n u_{ik}^m \|x_i - v_k\|^2}{n* min_{j,k} \|v_j - v_k\|^2}
\end{equation}

When the number of clusters increases,  the value of quality indices mechanically increases, too. Then, the important question is: how  useful is the addition of a new cluster? To answer this question, the most common solutions are penalization and the Elbow Rule \cite{cattell1966scree}. 

The first way to penalize a quality index is to multiply it by a quantity that diminishes the index when the number of clusters increases. In this case, the main difficulty is to choose the penalty. For instance, the penalized version of $V_{FRatio}$ is Calinski's $V_{FCH}$ \cite{calinski1974dendrite} (Equation~\ref{CH_eqn}; $0 \leq V_{FCH} \leq +\infty$; to be maximized), where the penalty is based on both the number of clusters and data points.

\begin{equation} \label{CH_eqn}
V_{FCH} = \frac{FB/(K-1)} {FW/(n-K)} = \frac {n-K}{K-1} \frac {FB}{FW}
\end{equation}

The second way to penalize a quality index is to evaluate index evolution relatively to the number of clusters, by considering the curve of the index' successive values. The most appropriate value of $K$ can be determined visually by help of the Elbow Rule or algebraic calculation \cite{dimitriadou2002examination}. 

To construct a visual determination of the Elbow Rule, $K$ is represented on the horizontal axis and the considered quality index on the vertical axis. Then, we look for the value of $K$ where there is a change in the curve's concavity. This change represents the optimal number of clusters $K$. To construct an algebraic determination, let $i_K$ being the index value for $K$ clusters. The variation of $i_K$ before $K$ and after $K$ are compared. In case of a positive Elbow, the second difference $min_K((i_{K+1} - i_K)-(i_K - i_{K-1}))$ is minimized. Yet, since the values before $K$ and after $K$ are used for calculation, the Elbow Rule can be applied to more than two clusters only.

Among all the above-stated quality indices, there is no single quality index that gives the best result for any dataset. Thus, there is room for a new quality index that is specifically tailored for fuzzy validation and helps the user choose the value of $K$.

\section{An Index Associated with a Visual Solution}
\label{sec:AnIndexAssociatedwithVisualSolution}

Building a new quality index, we first consider $FW$ to evaluate compactness and $FB$ to evaluate separability. We can choose to calculate either $FB - FW$, which is similar to $V_{FS}$ except for the sign, or $FB \div FW$, which is similar to $V_{FRatio}$. Unfortunately, $FI = FB + FW$ is not constant and $FB - FW \in [-FI, +FI]$. To take this particularity of fuzzy clustering into account, we propose to standardize $FB - FW$ by considering the \textit{Standardized Fuzzy Difference} $SFD = (FB - FW) \div FI$ instead. Then, $SFD \in [-1, +1]$. 

Adding a new cluster often improves clustering quality mechanically. Thus, many authors penalize the quality index with respect to $K$ (the smaller $n$ is, the greater the penalty), e.g., $V_{FCH}$ (Section~\ref{sec:RelatedWorks}). To obtain a penalized index, $SFD$ is first linearly transformed in an index belonging to $[0, 1]$, obtaining the \textit{Transformed Standardized Fuzzy Difference} $TSFD$ (Equation~\ref{TSFD_eqn}; $TSFD \in [0, 1]$; to be maximized). Finally, by penalizing $TSFD$ as $V_{FCH}$, we obtain the  \textit{Penalized Standardized Fuzzy Difference} $PSFD$ (Equation~\ref{PSFD_eqn}; $PSFD \in [0, (n-K)/(K-1)]$; to be maximized). 

\begin{equation} \label{TSFD_eqn}
TSFD = \frac{1+SFD}{2} = \frac{FB}{FI} 
\end{equation}

\begin{equation} \label{PSFD_eqn}
PSFD = TSFD * \frac {n-K}{K-1} = \frac{FB - FW}{FI} * \frac {n-K}{K-1}
\end{equation}

\begin{figure}[!hbt]
\begin{center}
\includegraphics[width=12cm,height=5cm]{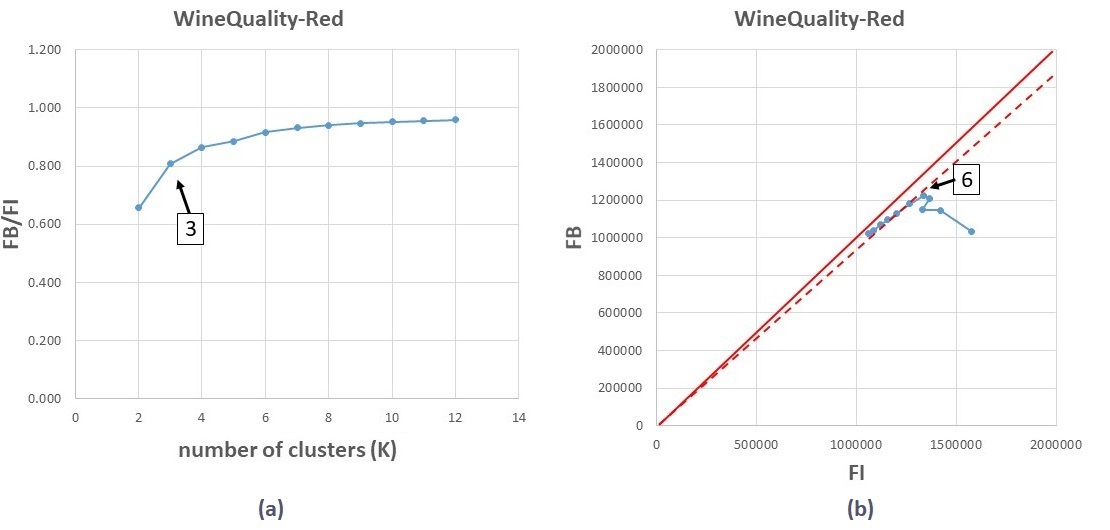}
\caption{Comparison of Elbow Rule (a) and \textit{Visual TSFD} (b) on the WineQuality-Red dataset (Table~\ref{tab:QualityIndicesResults})}
\label{figs:ExampleGraphSolution}
\end{center}
\end{figure}

Instead of penalizing the quality index, another solution is to visualize the search for the best number of clusters $K$. First solution is to apply the Elbow Rule to $TSFD$. $TSFD$ is plotted with respect to $K$ in Figure~\ref{figs:ExampleGraphSolution}(a). The drawback of this method is that the horizontal axis corresponds to an arithmetic scale of $K$ values, which is not satisfying. To fix this problem, we suggest to plot $FB$ with respect to $FI$, which we call \textit{Visual $TSFD$}. Our aim is not to give an automatic solution, but to help the user visually choose the most appropriate $K$ value. The visualization we propose is shown in Figure~\ref{figs:ExampleGraphSolution}(b), where the blue line plots $TSFD$ with respect to $K$, the full red line is the diagonal that corresponds to the best solutions ($FB = FI$) such that $TSFD=1$, and the dashed red line connects the origin to each point associated with $K$ values. The smaller the angle between the full red line and the dashed red line, the better is the solution. As the value of $K$ increases, the angle between the dashed red line and the diagonal decreases. Then, we choose the value of $K$ beyond which the decrease becomes negligible. This value is considered as the optimal number of clusters. For example, in Figure~\ref{figs:ExampleGraphSolution}(b), a first solution could be $K=4$, a better solution $K=6$, and it is not very interesting to consider $K > 6$. 

\section{Experimental Validation}
\label{sec:Experiments}

In this section, we compare our proposals $TSFD$, $PSFD$, \textit{Visual TSFD} and the use of the Elbow Rule to state-of-the-art clustering quality indices for FCM-like clustering algorithms, i.e., $V_{PC}$, $V_{CL}$, $V_{FCH}$, $V_{FS}$ and $V_{XB}$ (Section~\ref{sec:RelatedWorks}). 

In our experiments, the FCM algorithm is parameterized with its default settings: termination criterion $\epsilon = 0.0001$ and default fuzziness coefficient $m = 2$. All clustering quality indices are coded in Python version 2.7.4. 

\subsection{Datasets}
\label{subsec:ExperimentDatasets}

Quality indices are compared on ten real-life datasets (Table~\ref{tab:QualityIndicesResults}; IDs~1-10) from the UCI Machine Learning Repository\footnote{\url{http://archive.ics.uci.edu/ml/}} and seven artificial datasets (Table~\ref{tab:QualityIndicesResults}; IDs~11-17). In  real-life datasets, the true number of clusters is assimilated to the number of labels. Although using the number of labels as the number of clusters is debatable, it is acceptable if the set of descriptive variables explains the labels well. In artificial datasets, the number of clusters is known by construction. Moreover, we created new artificial datasets by introducing overlapping and noise to some of the existing datasets, such as E1071-3 \cite{meyer2017package}, Ruspini \cite{ruspini1970numerical} and E1071-5 \cite{meyer2017package} (Table~\ref{tab:QualityIndicesResults}; IDs 12-14). To create a new dataset, new data points are introduced, and each must be labeled. To obtain a dataset with overlapping, we modify the construction of the E1071 artificial datasets \cite{meyer2017package}. In the original datasets, there are three or five clusters of equal size (50). Cluster $i$ is generated according to a Gaussian distribution $N(i; 0.3)$. To increase overlapping in the three clusters while retaining the same cluster size, we only change the standard deviation from 0.3 to 0.4. Then, there is no labeling problem. To introduce noise in a dataset, we add in each cluster noisy points  generated by a Gaussian variable around each label gravity center. Noisy data are often generated by distributions with positive skewness. For example, in a two-dimensional dataset, for each label, we add points that are far away from the corresponding gravity center, especially on the right-hand side, which generally contains the most points. Then, we draw a random number $r$ between 0 and 1. If $r \leq 0.25$, the point is attributed to the left-hand side. Otherwise, the point is attributed to the right-hand side. This method helps obtain noisy data that are $^1/_4$ times smaller and $^3/_4$ times greater, respectively, than the expected value for the considered label. This process is applied to the Ruspini dataset \cite{ruspini1970numerical}.

\subsection{Experimental Results}
\label{subsec:ExperimentalResults}

In our experiments, all validation indices (Sections~\ref{sec:RelatedWorks} and~\ref{sec:AnIndexAssociatedwithVisualSolution}) are applied on all the datasets from Table~\ref{tab:QualityIndicesResults}. Moreover, since presenting all the results would take too much space, we retain only the best results for each index (even excluding $PSFD$). 

\begin{table}[!hbt]
\centering
\caption{Quality Indices Experiment Results with Different Datasets}
%\begin{tabular}{|c|c|c|c|c|c|c|c|c|c|c|c|}
\begin{tabular}{|c|c|c|c|c|c|c|c|c|c|c|c|}
\hline
ID & Datasets & \begin{tabular}[c]{@{}c@{}} \# of \\ data \\ points  \end{tabular} & \begin{tabular}[c]{@{}c@{}} \# of \\ clusters \end{tabular} & $V_{PC}$ & $V_{CL}$ & $FB$ & $V_{FCH}$ & $V_{FS}$ & $V_{XB}$ & \begin{tabular}[c]{@{}c@{}} Elbow \\ $V_{TSFD}$\end{tabular} & \begin{tabular}[c]{@{}c@{}} Visual \\ $V_{TSFD}$ \end{tabular} \\ \hline
1 & Wine & 178 & 3 & 2 & 2 & 8  & 12 & 8 & 2 & \textbf{3} & 5   \\ \hline
2 & Iris & 150 & 3  & 2 & 2  & \textbf{3} & \textbf{3} & \textbf{3} & 2 & \textbf{3}  & \textbf{3} \\ \hline
3 & Seeds & 210 & 3 & 2 & \textbf{3}  & \textbf{3} & \textbf{3} & \textbf{3} & 2 & \textbf{3}  & \textbf{3} \\ \hline
4 & Glass & 214 & 6 & 2 & 2  & 12  & 12 & 12 & 2 & 4 & 5,7 \\ \hline
5 & Vehicle & 846 & 4 & 2 & 2 & 2 & 2 & 5 & 2 & 3  & \textbf{4},5  \\ \hline
6 & Segmentation & 2310 & 7 & 2 & 4  & 4  & 4 & 12 & 12 & 3 & \textbf{7},8  \\ \hline
7 & \begin{tabular}[c]{@{}c@{}}Movement Libras \end{tabular} & 360 & 15 & 2 & 18 & 16 & 16 & 18 & 2 & 14 & 14,16 \\ \hline
8 & Ecoli & 336 & 8 & 2 & 3 & 3 & 3 & 12 & 3 & 3 & 3,7 \\ \hline
9 & Yeast & 1484 & 10 & 2 & 2 & 5  & 2 & 12 & 2 & 4 & 7,8 \\ \hline
10 & WineQuality-Red & 1599 & 6 & 2 & 2 & \textbf{6} & 7 & \textbf{6} & 2 & 3  & \textbf{6} \\ \hline
11 & Bensaid \cite{bensaid1996validity} & 49 & 3 & \textbf{3} & \textbf{3} & 9  & 11 & 11 & \textbf{3} & \textbf{3} & 5   \\ \hline
12 & E1071-3 \cite{meyer2017package} & 150 & 3  & \textbf{3} & \textbf{3} & \textbf{3} & \textbf{3} & \textbf{3} & \textbf{3} & \textbf{3}  & \textbf{3}  \\ \hline
13 & Ruspini \cite{ruspini1970numerical} & 75 & 4  & \textbf{4} & \textbf{4} & \textbf{4} & \textbf{4} & \textbf{4} & \textbf{4} & 3 & \textbf{4} \\ \hline
14 & E1071-5 \cite{meyer2017package} & 250 & 5  & 2 & \textbf{5} & 4 & \textbf{5} & \textbf{5} & 2  & 3 & \textbf{5} \\ \hline
15 & \begin{tabular}[c]{@{}c@{}} E1071-3- \\ overlapped \end{tabular} & 150 & 3 & 2 & \textbf{3} & \textbf{3} & 2 & \textbf{3} & 2 & \textbf{3} & \textbf{3} \\ \hline
16 & \begin{tabular}[c]{@{}c@{}} Ruspini\_noised \end{tabular} & 95 & 4 & \textbf{4} & 12 & \textbf{4} & \textbf{4} & \textbf{4} & \textbf{4} & \textbf{4} &  \textbf{4} \\ \hline
17 & \begin{tabular}[c]{@{}c@{}} E1071-5- \\ overlapped \end{tabular} & 250 & 5 & 2 & 2 & 4 & \textbf{5} & 4 & 2 & 3 & \textbf{5} \\ \hline
\multicolumn{4}{|c|}{\textbf{\#} \textbf{of} \textbf{wins for real-life datasets}} & 0 & 1 & 3 & 2 & 3 & 0 & 3 & \textbf{5} \\ \hline
\multicolumn{4}{|c|}{\textbf{\#} \textbf{of} \textbf{wins for artificial datasets}} & 4 & 5 & 4 & 5 & 5 & 4 & 4 & \textbf{6} \\ \hline
\multicolumn{4}{|c|}{ \textbf{Total} \textbf{\#} \textbf{of} \textbf{wins}} & 4 & 6 & 7 & 7 & 8 & 4 & 7 & \textbf{11} \\ \hline
\end{tabular}
\label{tab:QualityIndicesResults}
\end{table}

\begin{figure}[!hbt]
\begin{center}
\includegraphics[width=12cm,height=11.7cm]{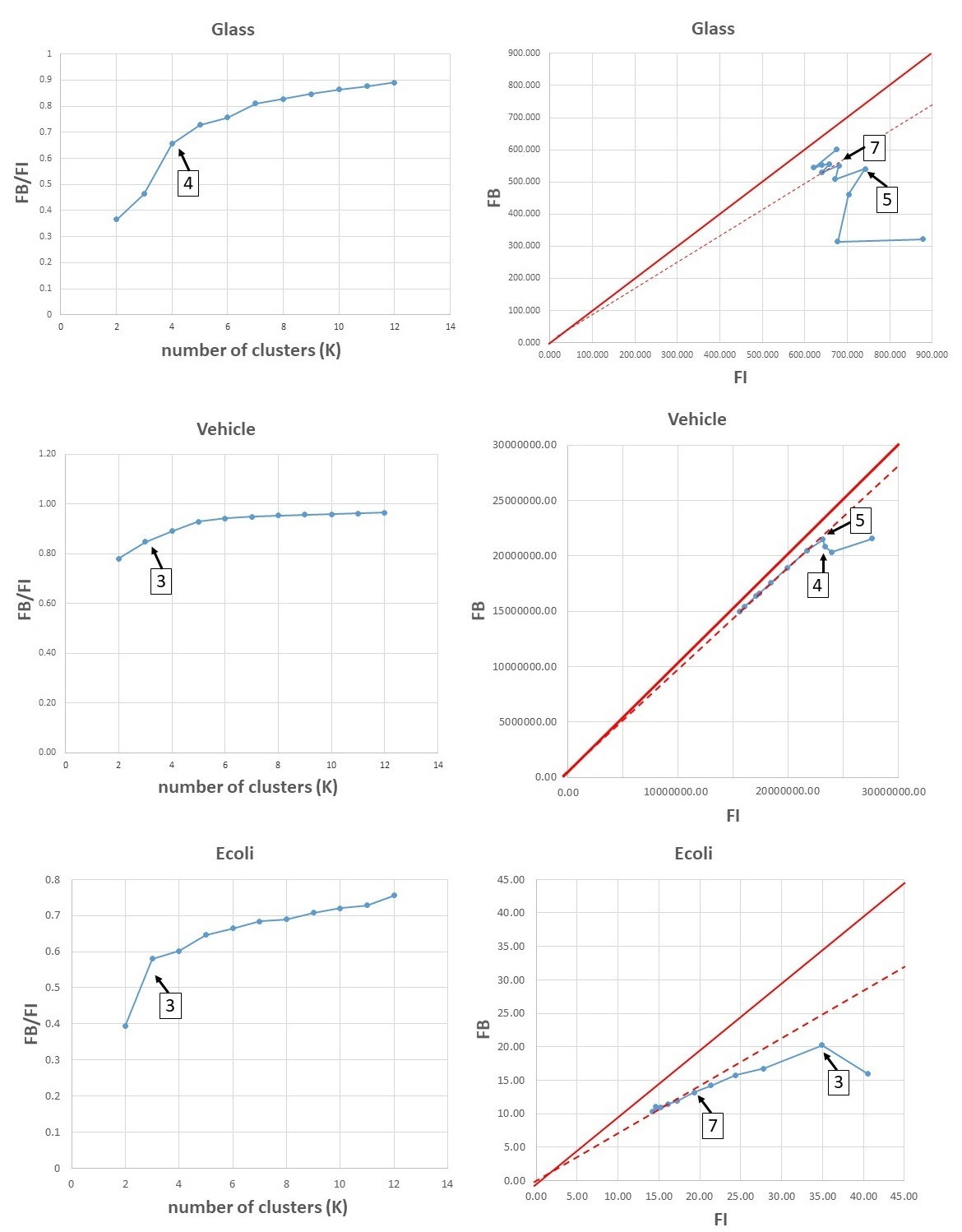}
\caption{Comparison of Elbow $TSFD$ and \textit{Visual TSFD} (1/2)}
\label{figs:GraphSolutionsResults-1}
\end{center}
\end{figure}

\begin{figure}[!hbt]
\begin{center}
\includegraphics[width=12cm,height=11.7cm]{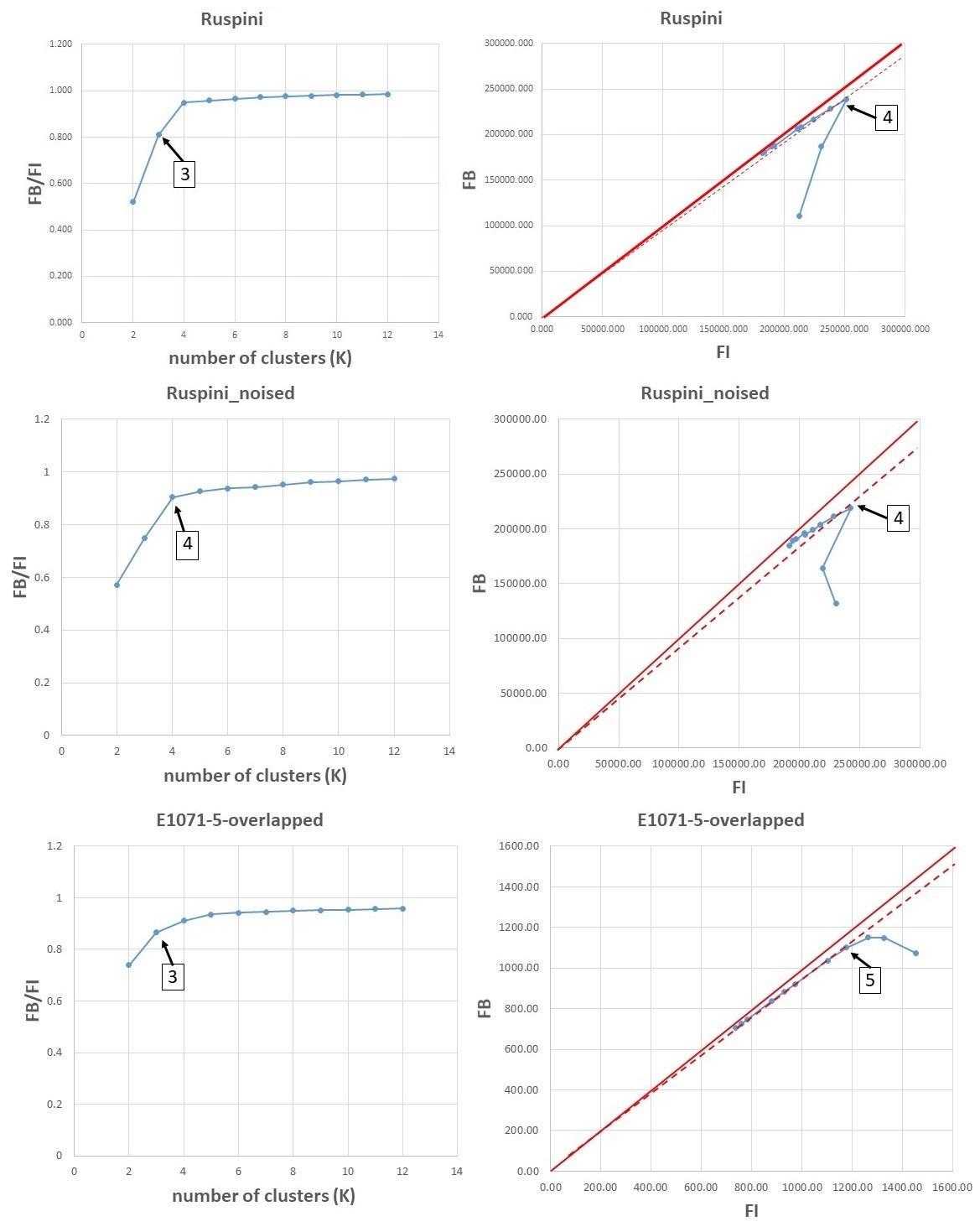}
\caption{Comparison of Elbow $TSFD$ and \textit{Visual TSFD} (2/2)}
\label{figs:GraphSolutionsResults-2}
\end{center}
\end{figure}

As shown in Table~\ref{tab:QualityIndicesResults}, it is more difficult to predict an appropriate number of clusters for real-life datasets than for artificial datasets. Considering all indices, the average rate of success is indeed 21\% in the case of real data, against 66\% in the case of artificial data. Whatever the type of data, \textit{Visual TSFD} outperforms the other indices, with 5 wins out of 10 in the case of real datasets, and 6 wins out of 7 in the case of artificial datasets. The worst results are obtained with $V_{PC}$ and $V_{XB}$ (0/10 and 4/7 wins each). The other indices achieve intermediary results. In addition, when the value given by \textit{Visual TSFD} is erroneous, it is quite close to the expected $K$, in contrast to $V_{FS}$, our closest competitor (Table~\ref{tab:QualityIndicesResults}; Wine, Glass, Segmentation, Ecoli and Bensaid). For example, the optimal number of clusters should be 6 for the Glass dataset. $V_{FS} = 12$, \textit{Visual TSFD}'s results are 5 and 7. Furthermore, we compare in Figures~\ref{figs:GraphSolutionsResults-1} and~\ref{figs:GraphSolutionsResults-2} \textit{Visual TSFD} and the plot obtained with the Elbow Rule (which is labeled Elbow $TSFD$) with respect to $K$, on a sample of both real-life and artificial datasets bearing different characteristics, i.e., Glass, Vehicle, Ecoli, Ruspini, Ruspini\_noised and E1071-5-overlapped (Table~\ref{tab:QualityIndicesResults}). As is clearly visible from Figures~\ref{figs:GraphSolutionsResults-1} and~\ref{figs:GraphSolutionsResults-2}, \textit{Visual TSFD} gives a better visual idea than Elbow $TSFD$. Elbow $TSFD$ indeed highlights $K$ values of 3 or 4, while the $TFSD$ blue plot systematically indicates larger $K$ values.  

Eventually, since our work aims at real-life datasets, there is no ground truth or golden standard for clustering analysis. In such a context, \textit{Visual TSFD} has the advantage of providing options to experts instead of outputting a single $K$ value. This makes our method more flexible that the existing ones in real-life scenarios.

\section{Conclusion and Perspectives}
\label{sec:Conclusion}

In this paper, we propose a novel quality index for FCM called \textit{Visual TSFD}, which provides an overview of fuzzy clustering with respect to the number of clusters. We compare \textit{Visual TSFD} to several clustering quality methods from the literature and experimentally show that it outperforms existing methods on various datasets. Furthermore, \textit{Visual TSFD} can also be used in the case of categorical data with Fuzzy K-Medoids \cite{park2009simple}. Thus, \textit{Visual TSFD} allows to deal with heterogeneous datasets, which makes our method a simple but noteworthy contribution, in our opinion. As a result, our next step is to design an ensemble fuzzy clustering method based on \textit{Visual TSFD} that would deal with both numerical and categorical data.

\section*{Acknowledgments} This project is supported by the Rh\^{o}ne Alpes Region's ARC~5: ``Cultures, Sciences, Soci\'et\'es et M\'ediations'' through A. \"Ozt\"urk's Ph.D. grant.

\bibliographystyle{splncs}
\bibliography{references}
\end{document}